\documentclass{article}

\usepackage[preprint]{automl}

\usepackage[utf8]{inputenc} % allow utf-8 input
\usepackage{authblk}
\usepackage[T1]{fontenc}    % use 8-bit T1 fonts
\usepackage{hyperref}       % hyperlinks
\usepackage{url}            % simple URL typesetting
\usepackage{booktabs}       % professional-quality tables
\usepackage{amsfonts}       % blackboard math symbols
\usepackage{nicefrac}       % compact symbols for 1/2, etc.
\usepackage{microtype}      % microtypography

\usepackage{cite}
\usepackage{graphics}
\usepackage{subfigure}
\usepackage{epsfig}
\DeclareGraphicsExtensions{.eps}

\title{An Auto-ML Framework Based on GBDT for Lifelong Learning}

\author{
  \textbf{Jinlong Chai\thanks{The authors contributed equally to this study.}, Jiangeng Chang, Yakun Zhao, Honggang Liu}\\
  Beijing University of Posts and Telecommunications, Central South University, Beijing University of Posts and Telecommunications, Inspur Electronic Information Industry Co.,Ltd \\
  \texttt{chaijinlong@bupt.edu.cn, 0109150105@csu.edu.cn, zhaoyakun@bupt.edu.cn, liuhonggang@inspur.com}
}

\begin{document}

\maketitle

\begin{abstract}
Automatic Machine Learning (Auto-ML) has attracted more and more attention in recent years, our work is to solve the problem of data drift, which means that the distribution of data will gradually change with the acquisition process, resulting in a worse performance of the auto-ML model. We construct our model based on GBDT, Incremental learning and full learning are used to handle with drift problem. Experiments show that our method performs well on the five data sets.  Which shows that our method can effectively solve the problem of data drift and has robust performance.
\end{abstract}

\section{Introduction}

% （第一段）简单介绍AutoML的概念。结尾几句话介绍一下比赛。【柴】
Over the last few years, machine learning (ML) has achieved impressive success, and more and more disciplines need to rely on it. However, current ML require human ML engineers help with pre-processing data, feature selection, model selection, parameter adjustment, etc. We hope that non-experts can experiment with ML to produce high-performing models. Automatic Machine Learning (Auto-ML) aims at achieving the automation of ML to free up developer’s time to focus on other aspects of tasks. The NeurIPS Auto-ML competition invites participants to design a model that can be self-trained, predicted and evaluated in a lifelong machine learning environment with limited resources and time.

% \section{Main Idea}
% （第四段）简要介绍一部分其他人的精彩观点（注意不要提别人的队伍）。重点描述我们的思想，注意是思想。【常】
The test of this competition will use multiple data sets which are anonymous, in consequence, it is difficult for us to build the corresponding professional features with
experience. By analyzing original methods, we found that constructing multidimensional features may be a viable approach, although it risks memory overflow. Therefore, we have constructed another automated feature method that first exploits the feature data with OrdinalEncoder, frequency, and logarithmic regression to extract the potential of the features. At the same time, Full-quantity learning is adopted to balance the distribution of data. In addition, we add the residuals from the previous data set to the next data set to speed up the fit of the trees. Through the above methods, it is possible to prevent more stress on the memory while mining data information as much as possible. Besides, we use the method of pre-training to greatly enhance the training speed. Moreover, we can ensure that our model can learn from more important data by adjusting the learning rate of the model.

% （第五段）柴金龙来写。【柴】
The rest of this article is organized as follows. Section II reviews the related work. Section III introduces the framework for lifelong learning. Section IV shows the performance of our framework. Finally, Section V summarizes the paper.

\section{Retlated Work}
% （第二段）关于autoML的最新研究。【赵】
% \section{Current research}
The development of Auto-ML has been rapid in recent years. [1] list newest strategies for evaluation and optimization. Heuristic search, derivative-free optimization, gradient descent and reinforcement learning are used to optimize its search process. Besides, Meta-learning and transfer learning are also playing a great role in advanced evaluation strategies. Algorithms and strategies have been devoted to getting a better performance with lower computation budgets, however, the phenomenon of concept drift often leads to the low practicability of the whole algorithm.

% （第三段）关于解决数据漂移的最新研究。【柴】
The problem of concept drift has received the attention of many researchers many years ago. H. Wang et al. proposed a general framework for mining concept drifting data streams using weighted ensemble classifiers [2]. In addition, the classifiers in the ensemble are judiciously weighted based on their expected classification accuracy on the test data under the time-evolving environment. In [3], M. Scholz and R. Klinkenberg proposed a boosting-like method to train a classifier ensemble from data streams that naturally adapts to concept drift. Moreover, it allows to quantify the drift in terms of its base learners. [4] compares the three methods of maintaining an adaptive time window, selecting representative training examples and weighted training examples.

\section{Framework based on GBDT for lifelong learning}
Here, we will introduce our main work in this competition. The technical framework will be introduced firstly. The following are the several parts of the whole model construction.
% (第一段）概况模型的思想。【宏刚】
\subsection{Framework design}
In this competition, the following questions were repeatedly considered when designing the framework model.

1. Algorithm versatility. Robust classification performance on different types of datasets

2. Lifelong learning. capability of adapting to changes in data distribution

3. Time budget and memory. reasonable allocation of program runtime and avoidance of memory overflow

For the above problems, the base framework is designed is as follows.
\begin{figure}[htbp]
    \centering
    \includegraphics[width=\textwidth]{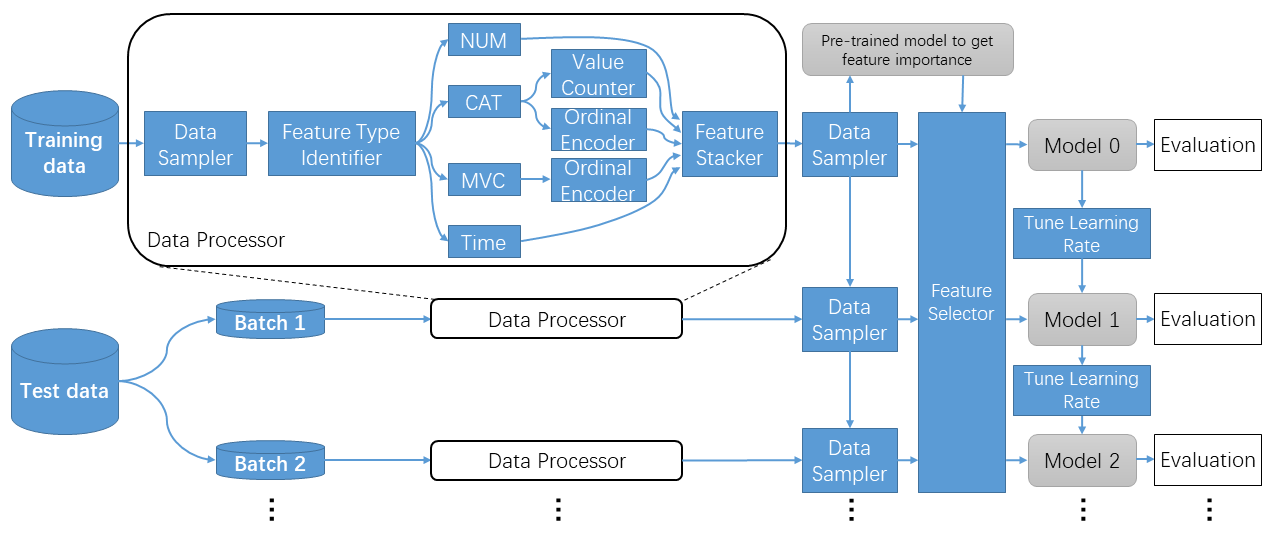}
    \caption{Base framework for lifelong machine learning.}
    \label{train_score}
\end{figure}

% （第二段...） 花几段分别介绍模型的不同模块，注意结合数据。【等图画出来再分，估计宏刚、常、赵负责】
% Data Processor部分【赵】
\subsection{Pre-processing}
For pre-processing step, we deal with data from two perspectives. On the one hand, nan value or null value should be filled to avoid computing error, on the other hand, data needs to be re-coded to prevent performance degradation due to unnecessary correlation introduced by OrdinalEncoder, we also coded the data with the frequency of the category. This guarantees both computational speed and performance. It is worth mentioning that we coding one feature for twice, once ordinal and once frequency. Both of these features are fed into the model. And before this step, necessary operations are needed to reduce the amount of data
% 采样和特征选择部分【宏刚】
\subsection{Sampling and feature selection}
Different datasets generate a large number of features according to the same data processor method, resulting in a sharp increase in memory and runtime. The same feature has different effects on the classification model of different datasets. Therefore, it is particularly important to remove redundant features. The pre-trained model method is used to select a certain proportion of features based on feature importance for model training and prediction. In addition, the sliding window of the data is constructed. The classification model is updated by dynamically updating the data to achieve the purpose of processing the concept drift problem.
% 学习率调节部分【常】
\subsection{Parameter adjustment}
We have developed a method of adjusting the learning rate based on the importance of data. In the face of CTR, we judge the more useful the data in the future based on the chronological order. Therefore, we continue to increase our initial learning rate by the time of different batches, the formula is:$$p_{t}=p_{t-1}+0.01*n_{t}$$

Here $p_{t}$ is the current initial learning compensation, $p_{t-1}$ is the initial learning step of the previous time, and $n_{t}$ is the current batch.
% （最后一段）介绍下模型的参数调节。【赵】

In addition to learning rate,there are hundreds of parameters in our model but most of them are well initialized, therefore, some of the parameters were chosen to be adjusted according to the data set.early-stopping was used instead of fixed num-iterations can ensure the optimal solution. Reg-alpha, reg-lambda, min-split-gain can effectively control the complexity of spanning trees.Considering the time constraints, we set these parameters more strictly to control the complexity.

\section{Experiments}
Combining the above work, we have done several experiments on the final results. The experimental results are as follows.
% （第一段）介绍数据集的基本情况。【赵】
\subsection{Dataset introduction}

The competition offered five data sets named AA, B, C, D, E to test the performance of the algorithm. Each officially divided into 10 batches, of which 5 batches have target solutions. Competitors are allowed to use five targeted batches to construct the algorithm, and the final evaluation is carried out on the whole 10 batches, taking the average score of each batch. The details about the data sets are as Table \ref{data_info}. Each dataset has 4 types of features including categorical feature, numerical feature, multi-value categorical feature and time feature.
%In the header of the tabel, Dataset stands for the name of the five datasets. Cat,Num,MVC and Time represent the number of category features, numerical features, multiple-valued features and time series features, respectively. Size means the size of datasets,
Used/Budget represent the actual consumption time and the official time budget. Last but not the least, our program are run into the ckcollab/codalab-legacy docker running Python 3.6 (from Anaconda).

\begin{table}[htbp]
  \caption{datasets introduction.}
  \label{data_info}
  \centering
  \begin{tabular}{llllllll}
    \toprule
    \multicolumn{4}{c}{Feature}                   \\
    \cmidrule(){2-6}
    Dataset & Cat & Num & MVC & Time & Total & Size & Used[s]/Budge[s] \\
    \midrule
    AA & 51 & 23 & 6 & 2 & 82 & ~10 Million & 2821/3600 \\
    B & 17 & 7 & 1 & 0 & 25 & ~1.9 Million & 216/600 \\
    C & 44 & 20 & 9 & 6 & 79 & ~2 Million & 717/1200 \\
    D & 17 & 54 & 1 & 4 & 76 & ~1.5 Million & 567/600 \\
    E & 25 & 6 & 1 & 2 & 34 & ~17 Million & 1633/1800 \\
    \bottomrule
  \end{tabular}
\end{table}

\subsection{Results on AutoML3 Datasets}

The datasets in the final phase of AutoML3 challenge were invisible, so we use the scores of feedback phase to evaluate the performance of our program. The datasets of feedback phase was introduced in previous section.

% （第二段）出两张图，分别分析。【柴】
% \begin{figure}[!t]
%   \centering
%   \fbox{\rule[-.5cm]{0cm}{0cm}
%   \includegraphics[width=0.9\textwidth]{pic/train.png}
%   \rule[-.5cm]{0cm}{0cm}}
%   \caption{Sample figure caption.}
% \end{figure}

% \begin{figure}[!t]
%   \centering
%   \fbox{\rule[-.5cm]{0cm}{0cm}
%   \includegraphics[width=0.9\textwidth]{pic/test.png}
%   \rule[-.5cm]{0cm}{0cm}}
%   \caption{Sample figure caption.}
% \end{figure}

\begin{figure}[htbp]
    \centering
    \begin{minipage}[t]{0.48\textwidth}
    \centering
    \includegraphics[width=7.3cm]{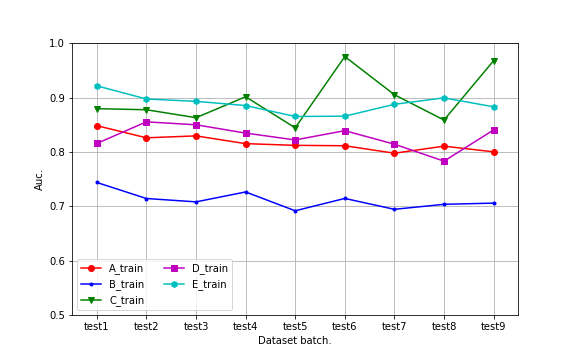}
    \caption{Train dataset.}
    \label{train_score}
    \end{minipage}
    \begin{minipage}[t]{0.48\textwidth}
    \centering
    \includegraphics[width=7.3cm]{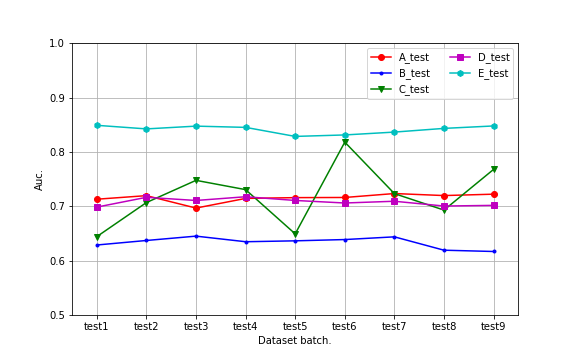}
    \caption{Test dataset.}
    \label{test_score}
    \end{minipage}
\end{figure}

The scores obtained in the feedback phase of the AutoML3 competition are shown in Figure \ref{train_score} and \ref{test_score}. Figure \ref{train_score} represents the scores of train datasets, which is the best scores with the best number of trees. It is not difficult to see that the datasets AA, B, D, E overcome the concept drift, and the model has a generalization performance. But the scores of the dataset C has fluctuated in the test5 and test6 batches. As shown in Figure \ref{test_score}, the same phenomenon occurs on the verification set.

% （第三段）总结一下仿真。【柴】
AutoML3 challenge has started in August 2018. In the feedback phase, our program ranked the 9th place. In addition, our program randed the 3th place in the final blind-test phase. Our programs are able to learn different datasets faster and have good generalization performance.

\section{Conclusion}
We designed the base framework to handle lifelong learning. This system has good algorithm versatility, good generalization performance when dealing with concept drift, and low time consumption. Although the current work ranks third in this competition, it still needs continuous improvement. In the future work, we will do more research on different data and learning tasks based on this framework.

% 总结

\section*{References}

\small

[1] Yao Q.M., Wang M.S., Hugo J.M., Isabelle G., Hu Y.Q., Li Y.F., Tu W.W., Yang Q.\ \& Yu Y. (2018)  Taking Human out of Learning Applications: A Survey on Automated Machine Learning.

[2] Wang H., Fan W., Yu P.S.\ \& Han J.\ (2003) Mining concept-drifting data streams using ensemble classifiers. {\it Proceedings of the ninth ACM SIGKDD international conference on Knowledge discovery and data mining} (pp. 226-235). ACM.

[3] Scholz M.\ \& Klinkenberg  R. (2007) Boosting classifiers for drifting concepts. {\it Intelligent Data Analysis} {\bf 11}(1):3-28.

[4] Klinkenberg R.\ (2004) Learning drifting concepts: Example selection vs. example weighting.  {\it Intelligent data analysis} {\bf 8}(3):281-300.

\end{document}